\documentclass{article}
\pdfoutput=1
\usepackage{kr}
\usepackage{times}
\usepackage{soul}
\usepackage{url}
\usepackage[hidelinks]{hyperref}
\usepackage[utf8]{inputenc}
\usepackage[small]{caption}
\usepackage{graphicx}
\usepackage{amsmath}
\usepackage{amsthm}
\usepackage{booktabs}
\usepackage{algorithm}
\usepackage{algorithmic}
\usepackage{amssymb}
\usepackage{array}
\usepackage{circuitikz}
\usepackage{color}
\usepackage{colortbl}
\usepackage{enumerate}
\usepackage{subcaption}
\usepackage{tikz}
\usepackage{xspace}
\usetikzlibrary{positioning}
\usetikzlibrary{calc}
\usetikzlibrary{arrows}
\usepackage{tangocolors}
\usepackage[nomessages]{fp} 

\newcommand{\alert}[1]{\textcolor{red!80!black}{#1}}

\usepackage{xstring}
\newcommand{\withproofs}{\newcommand{\usingproofs}{0}}
\newcommand{\Proof}[2][ ]{%
  \ifdefined\usingproofs%
    \IfStrEq{#1}{ }{%
      \begin{proof} #2 \end{proof}%
      }{%
        \begin{proof}[#1] #2 \end{proof}%
      }%
  \fi}

\newcommand{\citeay}[1]{\citeauthor{#1} (\citeyear{#1})}

\newcommand{\tup}[1]{\ensuremath{\langle #1 \rangle}}
\newcommand{\set}[1]{\ensuremath{\{ #1 \}}}
\newcommand{\Omit}[1]{}

\newcommand{\myeq}{\,{=}\,}

\newcommand{\domain}{\ensuremath{D}}
\newcommand{\instance}{\ensuremath{I}}
\newcommand{\goal}{\ensuremath{\mathit{Goal}}}
\newcommand{\initial}{\ensuremath{\mathit{Init}}}

\newcommand{\pre}{\ensuremath{\mathit{pre}}}
\newcommand{\eff}{\ensuremath{\mathit{eff}}}

\newcommand{\object}[1]{\text{\small\sf #1}}

\newcommand{\Dom}{\text{Dom}}

\newcommand{\iw}[1]{\ensuremath{\text{IW}(k)}\xspace}

\newcommand{\QT}{\ensuremath{\Q_{\mathcal{T}}}\xspace}

\newcommand{\reduct}[1]{\ensuremath{#1/\!\!\sim}\xspace}
\newcommand{\isoreduct}[1]{\ensuremath{#1/\!\!\iso}\xspace}
\newcommand{\eqX}[1]{\sim_{#1}}
\newcommand{\iso}{\eqX{iso}}

\newcommand{\SIM}[2]{#1\,{\sim}\,#2}
\newcommand{\ISO}[2]{#1\,{\iso}\,#2}

\newcommand{\nauty}{{\small\texttt{nauty}}\xspace}
\newcommand{\OG}{G} 

\newcommand{\states}{\ensuremath{S}}
\newcommand{\initialstate}{\ensuremath{s_I}}
\newcommand{\goalstates}{\ensuremath{G}}
\newcommand{\actions}{\ensuremath{\mathit{Act}}}
\newcommand{\successor}{f}
\newcommand{\Succ}{\text{Succ}\xspace}
\newcommand{\applicability}{\ensuremath{A}}

\newcommand{\A}{\mathfrak{A}}

\newcommand{\Q}{\ensuremath{\mathcal{Q}}}
\renewcommand{\S}{\ensuremath{\mathcal{S}}\xspace}

\newcommand{\colorclass}{\ensuremath{\mathcal{C}}}

\newcommand{\B}{\ensuremath{\mathfrak{B}}\xspace}

\newcommand{\C}{\ensuremath{\mathsf{C}}\xspace}

\newcommand{\hist}{\mathsf{Hist}}

\newtheorem{definition}{Definition}
\newtheorem{theorem}[definition]{Theorem}
\newtheorem{corollary}[definition]{Corollary}
\newtheorem{lemma}[definition]{Lemma}

\newtheorem*{example}{Example}

\title{Symmetries and Expressive Requirements for Learning General Policies}

\author{%
Dominik Drexler$^1$\and
Simon St\r{a}hlberg$^2$\and
Blai Bonet$^{3}$\and
Hector Geffner$^{2}$ \\
\affiliations
$^1$Link\"{o}ping University, Sweden\\
$^2$RWTH Aachen University, Germany\\
$^3$Universitat Pompeu Fabra, Spain\\
\emails
dominik.drexler@liu.se,
simon.stahlberg@gmail.com,\\
bonetblai@gmail.com,
hector.geffner@ml.rwth-aachen.de
}

\begin{document}
\withproofs
\maketitle

\begin{abstract}
  State symmetries play an important role in planning and generalized planning.
  In the first case, state symmetries can be used to reduce the size of the
  search; in the second, to reduce the size of the training set. In the case of
  general planning, however, it is also critical to distinguish non-symmetric
  states, i.e., states that represent non-isomorphic relational structures.
  However, while the language of first-order logic distinguishes non-symmetric
  states, the languages and architectures used to represent and learn general
  policies do not. In particular, recent approaches for learning general
  policies use state features derived from description logics or learned via
  graph neural networks (GNNs) that are known to be limited by the expressive
  power of $\C_2$, first-order logic with two variables and counting.
  In this work, we address the problem of detecting symmetries in planning and
  generalized planning and use the results to assess the expressive requirements
  for learning general policies over various planning domains. For this, we map
  planning states to plain graphs, run off-the-shelf  algorithms to determine
  whether two states are isomorphic with respect to the goal, and run coloring
  algorithms to determine if $\C_2$ features computed logically or via GNNs
  distinguish non-isomorphic states. Symmetry detection results in more effective
  learning, while the failure to detect non-symmetries prevents general policies
  from being learned at all in certain domains.
\end{abstract}

\section{Introduction}

Generalized planning is concerned with the problem of obtaining general action strategies for solving classes of instances drawn from
a common domain. A classical planning domain ensures that all instances share a structure given by a set of action schemas
and predicates. These general strategies, called also  general plans or policies, are learned
by considering a small set of training instances from the target class $\Q$
\cite{srivastava-et-al-aij2011,jimenez-et-al-ker2019,illanes-mcilraith-aaai2019,toyer-et-al-jair2020,yang-et-al-ijcai2022,srivastava-jair2022}.
General policies that solve the training instances are then expected to generalize to $\Q$.
In the symbolic setting, where the learning problem is formulated as a combinatorial optimization problem,
this generalization can often be established formally \cite{bonet-et-al-aaai2019,frances-et-al-aaai2021}.
In the deep learning setting, the algorithms scale up better but do not result in
policies that can be understood and proved to be correct \cite{stahlberg-et-al-kr2022,stahlberg-et-al-kr2023}.

The computational bottleneck of the symbolic approach 
is that it considers the complete state space of the training
instances, which becomes very large quickly. For example, in the Gripper domain, where the task is to move balls from one room to another,
the state space contains more than $2^n$ reachable states when the number of balls is $n$.
It turns out, however, that many pairs of states in the training set are \emph{symmetric},
meaning that a solution for one state implies a solution for the other.
This suggests that the number of states for the training set can be significantly reduced by considering
just one representative of each equivalent class of states.

Interestingly, state symmetries play a second important role in generalized planning.
Languages and neural architectures that lack the expressive power to distinguish pairs of states
that are \emph{not} symmetric may fail to represent general policies at all for certain domains.
In particular, recent approaches for learning general policies that use state features derived
from description logics or learned via graph neural networks (GNNs)
\cite{frances-et-al-aaai2021,stahlberg-et-al-arxiv2024} are known to be limited by the expressive power of $\C_2$,
first-order logic with two variables and counting \cite{barcelo-et-al-iclr2020,grohe:gnn}.

In this work, we address the problem of detecting symmetries in planning and generalized
planning and use the results for two different purposes: to assess the expressive requirements for
learning general policies over planning domains, which requires distinguishing non-symmetric states,
and to speed up learning, which involves grouping symmetric states together. 
For detecting symmetries, we map planning states to plain graphs, run off-the-shelf graph algorithms
to determine whether two states are isomorphic with respect to the goal,
and run coloring algorithms to determine if $\C_2$ features computed logically or via GNNs distinguish non-isomorphic states.
The expressive requirements and the performance gains are then evaluated experimentally.

The paper is organized as follows. After discussing related work, we review planning, generalized planning, and relational structures and graphs.
Then, we introduce faithful and uniform abstractions, look at the notion of isomorphic relational structures (states)
and the computation of such abstractions, carry out experiments, and draw conclusions.
\section{Related Work}

We discuss briefly three related research threads.

\smallskip\noindent\textbf{Symmetries.}
The detection of symmetries in planning has been used to  prune the search space~\cite{shleyfman-et-al-aaai2015}, to
define  heuristic functions~\cite{edelkamp-ecp2001,haslum-et-al-aaai2007,helmert-et-al-jacm2014,nissim-et-al-ijcai2011}, and
to transform the problem representation~\cite{riddle-et-al-icaps2016wshsdip}.
A common thread in  these  approaches, which contrasts with our approach,  is that actions are  explicitly considered
in the detection of symmetries \cite{pochter-et-al-aaai2011,sievers-et-al-icaps2019,sievers-et-al-icaps2017wshsdip-a}.

\Omit{
We show that as long as action schemas have a specific structure, explicit consideration is not necessary.
For example, the method proposed by~\citeay{pochter-et-al-aaai2011} uses off-the-shelf algorithms to compute automorphisms on a \textit{Problem Description Graph} (PDG).
Notably, this graph contains state variables and ground actions.
The PDG has been used in subsequent research~\cite{shleyfman-et-al-aaai2015}, including lifted versions that encode action schemas rather than ground actions~\cite{sievers-et-al-icaps2019}.
Other approaches use abstract representations to identify symmetries~\cite{sievers-et-al-icaps2017wshsdip-a}, where this representation includes actions.
It is worth noting that some symmetry breaking methods  consider both the initial state and the goal; however, we are only interested in whether two states are equivalent with respect to the goal.
}

\smallskip\noindent\textbf{General policies.}
The problem of learning general policies has a long history~\cite{khardon-aij1999,martin-geffner-ai2004,fern-et-al-jair2006,jimenez-et-al-ker2019}.
General, symbolic policies have been formulated in terms of logic~\cite{srivastava-et-al-aij2011,illanes-mcilraith-aaai2019},
regression~\cite{boutilier-et-al-ijcai2001,wang-et-al-jair2008,sanner-boutelier-aij2009},
and policy rules \cite{frances-et-al-aaai2021,drexler-et-al-icaps2022,yang-et-al-ijcai2022,srivastava-jair2023,silver-et-al-aaai2024}.
General policies have also been learned using deep learning methods
\cite{toyer-et-al-jair2020,bajpai-et-al-neurips2018,rivlin-et-al-icaps2020wsprl,stahlberg-et-al-icaps2022},
in many cases using graph neural networks or GNNs \cite{scarselli-et-al-ieeenn2009,gilmer-et-al-icml2017,hamilton-2020}.

\smallskip\noindent\textbf{Expressivity.}
Interestingly, the expressive limitations of  symbolic
methods relying on features derived  from the domain
predicates via description logic grammars \cite{bonet-et-al-aaai2019,frances-et-al-aaai2021}
and methods relying on GNNs \cite{stahlberg-et-al-kr2022,stahlberg-et-al-kr2023}
are similar. Such methods cannot  distinguish states (i.e., relational structures)
that cannot be distinguished by $\C_2$, first-order logic with two variables
and counting~\cite{barcelo-et-al-iclr2020,grohe:gnn},
or equivalently, by the Weisfeiler-Leman (1-WL)  coloring procedure \cite{cai-furer-immerman-combinatorica1992,morris-et-al-aaai2019,xu-et-al-iclr2019}.
The consequences of this limitation have been analyzed by \citeay{stahlberg-et-al-icaps2022}, and more recently by \citeay{horcik-sir-icaps2024}.
We will come back to this work in the discussion section.

\section{Background}

We review basic notions of planning, generalized planning, relational structures, and graphs.

\subsection{Classical Planning}

A \textbf{planning problem} is a pair $P\myeq\tup{\domain,\instance}$ where $\domain$
is a general first-order \emph{domain} containing a set of predicates (or relations) $R$,
each with given arity, and a set of action schemas of the form $\tup{\pre,\eff}$ where
$\pre$ is an arbitrary first-order formula and $\eff$ is an arbitrary effect,
and $\instance$ is specific instance
information that contains the set of objects $O$, and two sets of \emph{ground atoms,} $\initial$
and $\goal$, that describe the initial and goal situations, respectively.
The problem $P$ defines the state model $S_P^\circ\myeq\tup{\states,\initialstate,\goalstates,\actions,\applicability,\successor}$
where the states in $S$ are the truth valuations over the ground atoms, where each such valuation
is represented by the set of atoms true in the valuation, $\initialstate\myeq\initial$ is the initial state,
and $\goalstates=\set{s\in S \mid \goal\subseteq s}$  is the set of goal states.
The function $A$ maps states $s$ into the set $A(s)$ of ground actions from $Act$ that are applicable in $s$,
and the state transition function $f$ maps states $s$ and actions $a\in A(s)$ into the resulting state $s'\myeq f(s,a)$.

The \emph{unlabeled state model} for the problem $P$ is the tuple
$S_P\myeq\tup{\states,\initialstate,\goalstates,\Succ}$
where the actions are  compiled away, and  states have a set of possible successor states instead.
In this unlabeled model, the first  three components are those  for $S_P^\circ$, while
$\Succ=\set{(s,f(s,a)) \mid a\,{\in}\,A(s)}$ is the (unlabeled) successor relation.

A trajectory seeded at state $s_0$ in $P$ is a state sequence $s_0,s_1,\ldots,s_n$
such that $(s_i,s_{i+1})$ is in $\Succ$, $0\leq i<n$.
A state $s$ is reachable in $P$ if there is a trajectory seeded at the initial state $s_I$
that ends in $s$. For a reachable state $s$, a plan (resp.\ optimal plan) for $s$ is a
trajectory (resp.\ trajectory of minimum length) seeded at $s$ that ends in a goal state.
The length of an optimal plan for state $s$ is denoted by $V^*(s)$, and referred as
the optimal cost of state $s$.

\subsection{Generalized Planning}

A  \textbf{generalized planning problem} is a class $\Q$ of planning
problems $P$ for a common domain $\domain$ \cite{bonet-geffner-ijcai2018}.
A \textbf{general policy} $\pi$ for a class $\Q$ is a binary relation on states.
A state trajectory $s_0,s_1,\ldots,s_n$ is a $\pi$-trajectory seeded at state $s_0$
if $(s_i,s_{i+1})$ is a transition that is in both $P$ and $\pi$, for $0\leq i<n$.
We say that: (1)~$\pi$ solves state $s$ if each maximal $\pi$-trajectory
seeded at $s$ reaches a \textbf{goal state,} (2)~$\pi$ solves problem $P$ if it solves the
initial state of $P$, and (3)~$\pi$ solves class $\Q$ if it solves each problem $P$ in $\Q$.

In generalized planning, \emph{goals are encoded as part of the
state}  as follows. For each atom $p(\bar o)$ that appears in the goal condition $G$,
a new relational symbol $p_g$ of the same arity of $p$ is created.
Then, the initial situation $I$ is extended with the atoms \set{p_g(\bar o)\,{\mid}\,p(\bar o)\,{\in}\,G}
which are \textbf{static} and thus remain in every reachable state~\cite{martin-geffner-ai2004}.
Adding these ``goal atoms'' in the state  allows
general policies/sketches to take the specific goal of the
instance into account, so they may generalize not just to
instances with different numbers of objects and initial states,
but also to instances with different goals.

General policies are often represented in terms of \textbf{state features.}
A state feature $\phi$ for class $\Q$ is a function that maps
the reachable states $s$ for the problems in $\Q$ into values $\phi(s)$.
The feature $\phi$ is Boolean if its values are Boolean values, and numerical
if its values are non-negative integers.
If $\Phi$ is a set of features, $\Phi(s)$ denotes the vector $(\phi(s))_{\phi\in\Phi}$.

\subsection{States, Relational Structures, and Graphs}

A (planning) state defines a \textbf{relational structure} $\A^s$ with universe $U^s\,{=}\,O$
for the set of objects $O$ in $s$, and interpretations $R^s\,{\subseteq}\,(U^s)^k$ for each predicate
$R$ of arity $k$ in the planning domain $\domain$, where $\tup{o_1,o_2,\ldots,o_k}\,{\in}\,R^s$
iff $R(o_1,o_2,\ldots,o_k)$ is true in $s$.
The \emph{signature} of a relational structure $\A$ is the set of relational symbols in $\A$.
We assume fully relational structures that contain no functions nor constants (nullary functions).
This type of structures are adequate for planning problems described in PDDL.

While a planning state defines a relational structure,
relational structures can be encoded by graphs,
a mapping that we will use to test state equivalence.
Recall that a {directed graph}, or graph, is a pair $G = (V,E)$ where $V$ is the set
of vertices and $E\subseteq V^2$ is the set of edges.
An {undirected graph} is a directed graph $G$ where $E$ is symmetric;
i.e., $(v,w)\in E$ iff $(w,v)\in E$.
Two graphs $G=(V,E)$ and $G'=(V',E')$ are \textbf{isomorphic}, denoted by $G\simeq_g G'$,
if there is a bijection $f : V\rightarrow V'$ such that $(u,v)\,{\in}\,E$
iff $(f(u),f(v))\,{\in}\,E'$.

A \emph{vertex-colored graph} is a tuple $G=(V,E,\lambda)$ where $(V,E)$ is a
graph, and $\lambda:V\rightarrow \colorclass$
maps vertices to the colors in $\colorclass$.
Two vertex-colored graphs $G=(V,E,\lambda)$ and $G'=(V',E',\lambda')$  are \textbf{isomorphic},
denoted as $G\simeq_g G'$, iff there is a \emph{color preserving isomorphism}
$f$ from $G$ to $G'$, i.e., $\lambda(v) = \lambda'(f(v))$ for $v\in V$.
If the graphs $G$ and $G'$ are isomorphic via the bijection $f$, we write
$f:G\rightarrow G'$.

\section{Abstractions}

We formalize  first the abstraction induced by an
\textbf{equivalence relation} $\sim$:

\begin{definition}[Abstraction]
  \label{def:abstraction}
  Let $\Q$ be a class of problems, let $\sim$ be an equivalence relation
  on the reachable states of the problems in $\Q$, and let $P$ be a problem
  in $\Q$ with unlabeled state model $S_P\myeq\tup{S,s_I,G,\Succ}$.
  The \textbf{abstraction} of $P$ induced by $\sim$, denoted by $\reduct{P}$,
  is the \textbf{unlabeled state model}
  $\tilde{S}_P\myeq\tup{\tilde{S},[s_I],\tilde{G},\widetilde{\Succ}}$ where
  \begin{enumerate}
    \item $\tilde{S}\,{\doteq}\,\set{[s] \mid s\in S}$ is the set of equivalence classes for $P$,
    \item $[s_I]$ is the equivalence class for initial state $s_I$ of $P$,
    \item $\tilde{G}\,{\doteq}\,\set{[s] \mid s\in G}$ is the set of goal classes, and
    \item $\widetilde{\Succ}\,{\doteq}\,\set{([s],[s']) \mid (s,s') \in \Succ}$.
  \end{enumerate}
  The abstraction $\reduct{\Q}$ is the class of abstractions $\tilde{S}_P$
  for the problems $P$ in $\Q$.
\end{definition}

The successor relation in $\tilde{S}_P$ is the
\emph{existential quantification} of the successor relation in $S_P$
where $([s],[s'])\in\widetilde{\Succ}$ iff there is a transition $(t,t')$
in $\Succ$ such that $\SIM{s}{t}$ and $\SIM{s'}{t'}$.  In particular, the
transition $(s,s')$ may not exist in $P$.
Hence, generalized plans that solve the abstraction $\tilde{S}_P$ do not
necessarily solve $P$. In the following, we write $\SIM{(s,s')}{(t,t')}$
to denote $\SIM{s}{t}$ and $\SIM{s'}{t'}$.

\begin{definition}[Faithful Abstractions]
  \label{def:faithful}
  Let $\Q$ be a class of problems, and let $\sim$ be an equivalence
  relation on the reachable states in $\Q$. The abstraction $\reduct{\Q}$
  is \textbf{faithful} iff
  \begin{enumerate}[1.]
    \item for any $P$ in $\Q$, any reachable transition $(s,s')$ in $P$,
      and any reachable state $t$ in $P$ with $\SIM{t}{s}$, there is a
      transition $(t,t')$ in $P$ such that $\SIM{(s,s')}{(t,t')}$, and
    \item if $\SIM{s}{t}$ for reachable states $s$ and $t$ in $P$,
      then $s$ is a goal state iff $t$ is a goal state.
  \end{enumerate}
\end{definition}

If the abstraction $\reduct{\Q}$ is faithful, the binary relation
that associates states $s$ in $\Q$ with their equivalence classes $[s]$ in
\reduct{\Q} is a \textbf{bisimulation} between the corresponding unlabeled
transition systems \cite{sangiori:bisimulation}. Indeed,

\begin{theorem}[Bisimulation]
  \label{thm:faithful-abstraction}
  Let $\reduct{\Q}$ be a faithful abstraction, and let $P$ be a problem
  in $\Q$. Then,
  1)~if $s_0,s_1,\ldots,s_n$ is a trajectory in $S_P$, then
  $[s_0],[s_1],\ldots,[s_n]$ is a trajectory in $\tilde{S}_P$, and
  2)~if $[s_0],[s_1],\ldots,[s_n]$ is a trajectory in $\tilde{S}_P$,
  for each $s'_0$ in $[s_0]$, there is trajectory $s'_0,s'_1,\ldots,s'_n$
  in $S_P$ with $\SIM{s'_i}{s_i}$ for $0\leq i\leq n$.
\end{theorem}
\Proof{
  The first claim is direct by the definition of $\tilde{S}_P$.
  For the second, notice that $([s_i],[s_{i+1}])$ in $\widetilde{\Succ}$
  implies there is a transition $(s''_i,s''_{i+1})$ with
  $\SIM{(s_i,s_{i+1})}{(s''_i,s''_{i+1})}$,
  for $0\,{\leq}\,i\,{<}\,n$.
  We construct the required trajectory in $S_P$ inductively.
  By faithfulness, there is $s'_1$ such that $(s'_0,s'_1)$ is in \Succ
  and $\SIM{s'_1}{s''_1}$. Hence, $\SIM{s'_1}{s_1}$.
  After constructing $s'_0,s'_1,\ldots,s'_k$,
  we have $\SIM{s'_k}{s_k}$. By faithfulness, there is transition $(s'_k,s'_{k+1})$
  with $\SIM{s'_{k+1}}{s''_{k+1}}$. Thus, $\SIM{s'_{k+1}}{s_{k+1}}$, and the
  trajectory can be extended with $s'_{k+1}$.
}

\begin{corollary}
  Let $\reduct{\Q}$ be a faithful abstraction, and let $P$ be a
  problem in $\Q$. If $s$ and $t$ are reachable states in $P$
  with $\SIM{s}{t}$, then $V^*(s)=V^*(t)$.
\end{corollary}

Faithfulness allows us to work with the abstraction, but
it does not take into account the form of the policy $\pi$.
Namely, it can be the case that a transition $(s,s')$ in $P$ belongs
to $\pi$ but not a transition $(t,t')$ with $\SIM{(t,t')}{(s,s')}$.
This will not happen, however, for the large class of \emph{uniform policies}:

\begin{definition}[Uniform Policies]
  \label{def:uniform}
  Let $\reduct{\Q}$ be an abstraction, and let $\Pi$ be a class of policies
  for $\Q$. A policy $\pi$ in $\Pi$ is \textbf{uniform over $\reduct{\Q}$}
  iff for any problem $P$ in $\Q$, and any pair $(s,s')$ of reachable states in $P$,
  if $(t,t')$ is a pair of reachable states in $P$ such that $\SIM{(s,s')}{(t,t')}$,
  then $(s,s')$ is in $\pi$ iff $(t,t')$ is in $\pi$.
  The class $\Pi$ of policies is uniform over $\reduct{\Q}$  if each policy $\pi$ in $\Pi$ is so.
\end{definition}

A uniform policy $\pi$ over a faithful abstraction $\reduct{\Q}$
generates \textbf{well-defined trajectories} $[s_0],[s_1],[s_2],\ldots$
on the abstraction.
Let us say that the transition $([s],[s'])$ belongs to $\pi$ if $(s,s')$
belongs to $\pi$. By uniformity, if $t$ and $t'$ are reachable states
such that $\SIM{(s,s')}{(t,t')}$, then $(t,t')\in\pi$.
Hence, we can lift the notions of solvability to define when a policy
$\pi$ solves the abstraction \reduct{\Q}.
We have

\begin{theorem}[Solvability]
  \label{thm:solvability}
  Let $\reduct{\Q}$ be a faithful abstraction, and let $\Pi$ be a uniform
  class of policies for $\reduct{\Q}$. Then, for any policy $\pi$ in $\Pi$: $\pi$ solves $\Q$ iff $\pi$ solves
  $\reduct{\Q}$.
\end{theorem}
\Proof{
  Let us assume that $\pi$ solves $\Q$, and suppose it does not solve $\reduct{\Q}$.
  That is, there is a $P$ in $\Q$ with initial state $s_0$, and \textbf{maximal}
  trajectory $[s_0],[s_1],\ldots,[s_n]$ seeded at the initial class $[s_0]$ of $\tilde{S}_P$
  that is not goal reaching.
  By Theorem~\ref{thm:faithful-abstraction}, there is a trajectory $s'_0,s'_1,\ldots,s'_n$
  in $P$ such that $\SIM{s'_i}{s_i}$, for $0\leq i\leq n$.
  By faithfulness and uniformity, such a trajectory is a maximal $\pi$-trajectory.
  On the other hand, the state $s'_n$ cannot be a goal state since $[s_n]$
  is not a goal state.
  Hence, $\pi$ cannot solve $\Q$, which contradicts the assumption.
  The other direction is shown similarly.
}

In the next section, we define an equivalence relation over states
that yields faithful abstractions and uniform policies, and which
thus benefits from Theorem~\ref{thm:solvability}.


\section{Isomorphic Relational Structures (States)}

As planning states are relational structures, it is natural to
deem two states as equivalent when their relational structures
are \emph{isomorphic,} defined as follows:

\begin{definition}[Isomorphic Structures]
  \label{def:isomorphism}
  Two relational structures $\A$ and $\B$, over a \textbf{common} universe $U$
  and \textbf{common} signature (without constants), are \textbf{isomorphic},
  written as $\A\,{\simeq}\,\B$, iff there is a permutation $\sigma$ on $U$
  such that for each relation $R$ of arity $k$,
  $R^\B\myeq\set{ \sigma(\bar u)\,{\mid}\,\bar u\,{\in}\,R^\A }$,
  where $\sigma(\bar u)$ for tuple $\bar u\,{=}\,\tup{u_1,u_2,\ldots,u_k}$
  is the tuple $\tup{\sigma(u_1),\sigma(u_2),\ldots,\sigma(u_k)}$.
  We say that $\sigma$ maps $\A$ into $\B$, and write $\sigma:\A\rightarrow\B$.
\end{definition}

Isomorphic structures satisfy the same set of sentences and the same set of
formulas under suitable permutations. The following is a standard result.

\begin{lemma}
  \label{lemma}
  Let $\A$ and $\B$ be two relational structures, and let $\varphi(\bar x)$
  be a first-order formula whose free variables are among the ones in $\bar x$.
  If $\sigma: \A\rightarrow \B$, then for any tuple $\bar u$ of objects
  of the same length as $\bar x$, $\A\vDash\varphi(\bar u)$ iff $\B\vDash\varphi(\sigma(\bar u))$.
  In particular, if $\varphi$ is a sentence (i.e., it has no free variables),
  $\A\vDash\varphi$ iff $\B\vDash\varphi$.
\end{lemma}

In the STRIPS setting where classes $\Q$ consist of problems over
a common domain, isomorphism-based equivalence of states yields
faithful abstractions:

\begin{theorem}[Isomorphism-Based Equivalence]
  \label{thm:isomorphism}
  Let $\Q$ be a class of STRIPS problems over domain $D$.
  If $\iso$ is the equivalence relation on the reachable states in $\Q$
  such that $\ISO{s}{t}$ iff $\A^s\,{\simeq}\,\A^t$, then $\isoreduct{\Q}$
  is a \textbf{faithful abstraction.}
\end{theorem}
\Proof[Proof (sketch)]{
  Let $P$ be a problem in $\Q$, let $(s,s')$ be a reachable transition in $P$,
  and let $t$ be a reachable state in $P$ with $\ISO{t}{s}$. We need to show
  that there is a transition $(t,t')$ in $P$ with $\ISO{t'}{s'}$.
  By assumption, $\sigma:\A^s\rightarrow\A^t$ for some permutation $\sigma$,
  and there is a ground action $a(\bar o)$ with $s'\,{=}\,\successor(s,a(\bar o))$.
  In particular, $\A^s\vDash\pre(\bar o)$ and thus, by Lemma~\ref{lemma},
  $\A^t\vDash\pre(\sigma(\bar o))$ (i.e.\ the ground action $a(\sigma(\bar o))$
  is applicable in $t$).
  It is not hard to show that $\ISO{t'}{s'}$ for $t'\,{=}\,\successor(t,a(\sigma(\bar o)))$.

  Finally, to show the second condition in Definition~\ref{def:faithful}, let $P$ be a
  problem in $\Q$. As the states in $P$ are assumed to contain the goal atoms
  for the problem, the \textbf{sentence}
  $\varphi_g=\bigwedge_p \forall\bar x\bigl[ p_g(\bar x) \rightarrow p(\bar x) \bigr]$,
  where the conjunction is over all predicates $p$ in $D$, $p_g$ is the goal
  predicate for $p$, and the size of $\bar x$ is the arity of $p$, determines
  whether a state $s$ in $P$ is a goal state; i.e., $s$ is a goal state iff
  $\A^s\vDash \varphi_g$.
  Hence, if $s$ and $t$ are reachable states in $P$ such that $\ISO{s}{t}$, then
  $\A^s\vDash\varphi_g$ iff $\A^t\vDash\varphi_g$; i.e., $s$ is a goal state
  iff $t$ is a goal state.
}

\begin{figure}[t]
  \centering
  \resizebox{.99\linewidth}{!}{
  \begin{tikzpicture}[>=latex,
                      abstractState/.style={rectangle, rounded corners=2mm, draw, minimum width=25mm},
                      dummy/.style={},
                      edgeStyle/.style={thick,-latex}]
    \node[abstractState, align=left] (s0) {
      $\begin{array}{l}
         \#A = n \,, \quad\ \\
         \#G = 0 \,,\\
         L = \object{A}
       \end{array}$
    };

    \node[abstractState, right=2cm of s0] (s1) {
      $\begin{array}{l}
         \#A = n-1 \,,\\
         \#G = 1 \,,\\
         L = \object{A}
       \end{array}$
    };
    \node[abstractState, right=2cm of s1] (s2) {
      $\begin{array}{l}
         \#A = n-2 \,,\\
         \#G = 2 \,,\\
         L = \object{A}
       \end{array}$
    };
    \node[abstractState, below=1cm of s0] (s3) {
      $\begin{array}{l}
         \#A = n-1 \,,\\
         \#G = 1 \,,\\
         L = \object{B}
       \end{array}$
    };
    \node[abstractState, below=1cm of s2] (s4) {
      $\begin{array}{l}
         \#A = n-2 \,,\\
         \#G = 2 \,,\\
         L = \object{B}
       \end{array}$
    };
    \node[abstractState, below=1cm of s3] (s5) {
      $\begin{array}{l}
         \#A = n-1 \,,\\
         \#G = 0 \,,\\
         L = \object{B}
       \end{array}$
    };
    \node[abstractState, below=1cm of s4] (s6) {
      $\begin{array}{l}
         \#A = n-2 \,,\\
         \#G = 1 \,,\\
         L = \object{B}
       \end{array}$
    };
    \node[abstractState, below=1cm of s6] (s7) {
      $\begin{array}{l}
         \#A = n-2 \,,\\
         \#G = 0 \,,\\
         L = \object{B}
       \end{array}$
    };
    \node[abstractState, left=2cm of s6] (s8) {
      $\begin{array}{l}
         \#A = n-2 \,,\\
         \#G = 1 \,,\\
         L = \object{A}
       \end{array}$
    };
    \node[abstractState, below=1cm of s8] (s9) {
      $\begin{array}{l}
         \#A = n-2 \,,\\
         \#G = 0 \,,\\
         L = \object{A}
       \end{array}$
    };

    \draw[edgeStyle]   (s0) edge[transform canvas={yshift=5}, bend left=10] node[above] {$\mathit{pick}$} (s1); 
    \draw[edgeStyle]   (s1) edge[transform canvas={yshift=7}, bend left=10] node[below] {$\mathit{drop}$} (s0); 

    \draw[edgeStyle]   (s1) edge[transform canvas={yshift=5}, bend left=10] node[above] {$\mathit{pick}$} (s2); 
    \draw[edgeStyle]   (s2) edge[transform canvas={yshift=7}, bend left=10] node[below] {$\mathit{drop}$} (s1); 

    \draw[edgeStyle]   (s1) edge[transform canvas={xshift=0}, bend left=10] node[right] {$\mathit{move}$} (s3); 
    \draw[edgeStyle]   (s3) edge[transform canvas={xshift= 0}, bend left= 8] node[left] {$\mathit{move}$} (s1); 

    \draw[edgeStyle]   (s2) edge[transform canvas={xshift=2}, bend left=10] node[right] {$\mathit{move}$} (s4); 
    \draw[edgeStyle]   (s4) edge[transform canvas={xshift=-2}, bend left=10] node[left] {$\mathit{move}$} (s2); 

    \draw[edgeStyle]   (s3) edge[transform canvas={xshift=2}, bend left=10] node[right] {$\mathit{drop}$} (s5); 
    \draw[edgeStyle]   (s5) edge[transform canvas={xshift=-2}, bend left=10] node[left] {$\mathit{pick}$} (s3); 

    \draw[edgeStyle]   (s4) edge[transform canvas={xshift=2}, bend left=10] node[right] {$\mathit{drop}$} (s6); 
    \draw[edgeStyle]   (s6) edge[transform canvas={xshift=-2}, bend left=10] node[left] {$\mathit{pick}$} (s4); 

    \draw[edgeStyle]   (s6) edge[transform canvas={xshift=2}, bend left=10] node[right] {$\mathit{drop}$} (s7); 
    \draw[edgeStyle]   (s7) edge[transform canvas={xshift=-2}, bend left=10] node[left] {$\mathit{pick}$} (s6); 

    \draw[edgeStyle]   (s8) edge[transform canvas={yshift=5}, bend left=10] node[above] {$\mathit{move}$} (s6);
    \draw[edgeStyle]   (s6) edge[transform canvas={yshift=7}, bend left=10] node[below] {$\mathit{move}$} (s8);
    \draw[edgeStyle]   (s9) edge[transform canvas={yshift=5}, bend left=10] node[above] {$\mathit{move}$} (s7);
    \draw[edgeStyle]   (s7) edge[transform canvas={yshift=7}, bend left=10] node[below] {$\mathit{move}$} (s9);

  \end{tikzpicture}
  }
  \caption{Fragment of the state model $\tilde{S}_P$ for a Gripper instance with $n$ balls.
    Each equivalence class is identified by the number of balls at room \object{A} ($\#A$), the number of balls being held ($\#G$),
    and the position of the robot ($L$).
    For better understanding, we label transition with the action schemas that induce them.
    The abstraction contains $6n$ abstract states (see text).
  }
  \label{fig:abstraction}
\end{figure}
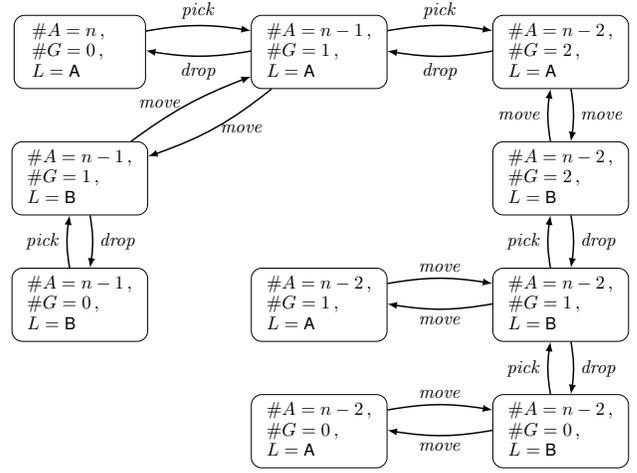

\begin{example}
  Let us consider the Gripper domain, where the goal is to move \textbf{all balls}
  from room \object{A} to room \object{B} with a robot.
  The robot has two grippers, it can move between the rooms, and
  it can pick and drop balls with any of the grippers.
  As the goal is for all balls to be in room \object{B}, 
  two states are equivalent if both have the same number
  of balls in each room, and the robot is in the same room in each state.

  If $P$ is an instance with $n$ balls, the number of non-isomorphic states
  is $6n=2[(n+1)+n+(n-1)]$: for each of the two possible positions of the
  robot, there are $n+1$ states with no ball being held, $n$ states with one
  ball being held, and $n-1$ states with two balls being held.
  On the other hand, the (plain) state space contains an exponential number of
  states: when no ball is being held, for example, each ball and the robot can
  be in either room, for a total of $2^{n+1}$ states.
  Thus, abstractions for Gripper are exponentially smaller.
  Figure~\ref{fig:abstraction} shows a fragment of the state model
  $\tilde{S}_P$ for the abstraction of $P$, where each ``abstract state''
  is represented with the features
  $\Phi\myeq\set{\#A,\#G,L}$ where $\#A$ counts the number of balls in room
  \object{A}, $\#G$ counts the number of balls being held, and $L$
  is the position of the robot, either \object{A} or \object{B}.
  The number of balls in room \object{B} is determined by
  the features $\#A$ and $\#G$. \qed 
\end{example}

The general policies $\pi$ defined in terms of rules \cite{bonet-et-al-aaai2019},
and GNNs \cite{stahlberg-et-al-icaps2022} are uniform for the abstraction $\isoreduct{\Q}$,
and hence, $\pi$  solves $\Q$ iff $\pi$ solves  $\isoreduct{\Q}$.
To see this, let us say that a policy  $\pi$ is \textbf{function-based} if there is a  function
$f$ that maps reachable states in $\Q$ into a domain $\Dom_f$ such that to determine whether
a state pair $(s,s')$ is in $\pi$, it is sufficient to look at the pair of values $(f(s),f(s'))$.
If the function $f$ is invariant under $\iso$, any policy $\pi$ that is based on $f$ is
uniform for \isoreduct{\Q}.
Likewise, policies that select pairs $(s,s')$ by looking at the set
$\set{(f(s),f(s'')) \mid (s,s'') \in \Succ}$, 
like policies that choose pairs $(s,s')$ that greedily minimize the value $f(s')$ over
successor states $s'$, are also uniform for \isoreduct{\Q} if $f$ is invariant.
Hence, we say that $\pi$ is an \textbf{invariant function-based} policy if $\pi$ is
based on a function $f$ that is invariant under $\iso$.
For such policies, Theorem~\ref{thm:solvability} implies:

\Omit{
The \textbf{rule-based policies} used in several symbolic approaches, e.g.\ 
are function-based policies,  as they select transitions by just looking at the pairs
$(\Phi(s),\Phi(s'))$ for a fixed set $\Phi$ of state features.
Likewise, \textbf{GNN-based policies} \cite{stahlberg-et-al-icaps2022} that
select transitions $(s,s')$ such that $f(s')\leq f(s)-1$, where $(f(s),f(s'))$
is the output value of the net on input $(s,s')$, are also function-based
policies.
On the other hand, policies whose selection $(s,s')$ depend on the path
leading to the state $s$ are not function-based policies.
}

\begin{theorem}[Main]
  \label{thm:function-based}
  Let $\Q$ be a class of STRIPS problems, and let $\pi$ be an
  invariant function-based policy for $\Q$.
  Then, $\pi$ solves $\Q$ iff $\pi$ solves \isoreduct{\Q}.
\end{theorem}
\Proof{
  Direct from Theorem~\ref{thm:solvability} as \isoreduct{\Q} is a faithful
  abstraction, by Theorem~\ref{thm:isomorphism}, and $\pi$ is uniform
  for \isoreduct{\Q}.
}
\section{Computing The Abstraction}

Checking $\iso$ on two reachable states can be reduced to a graph-isomorphism
test on vertex-colored graphs. These graphs, that we call \emph{object graphs},
encode relational structures as vertex-colored undirected graphs.
On the theoretical side, the exact complexity of graph isomorphism is
still unknown, but it can be tested in quasi-polynomial time~\cite{babai-stoc2016}.
However, in practice, the test can be performed efficiently~\cite{mckay-piperno-jsc2014};
see discussion in \citeauthor{babai-stoc2016} (\citeyear[page 83]{babai-stoc2016}).
Indeed, we use \nauty \cite{mckay-piperno-jsc2014} to compute \textbf{canonical representations}
(i.e.\ isomorphism-invariant representations) of graphs, that we apply to the object
graphs associated with states.
\nauty is a state-of-the-art tool that applies Color Refinement,
recursively, using a technique called vertex individualization.

\begin{definition}[Object Graphs]
  \label{def:graph}
  Let $\A$ be a relational structure with universe $U$, and relational symbols $R_i$,
  each of arity $k_i$, $0\leq i<n$.
  The \textbf{object graph} for $\A$ is the \textbf{vertex-colored undirected}
  graph $\OG(\A)=(V,E,\lambda)$ where the set $V$ of vertices consists of
  \begin{enumerate}
    \item vertices $v\myeq\tup{u}$ with color $\lambda(v)\myeq\bot$ for $u\,{\in}\,U$, and
    \item vertices $v\myeq\tup{R_i,j,\bar u}$ with color $\lambda(v)\myeq\tup{R_i,j}$ for
      each relation $R_i$, $1\leq j\leq k_i$, and tuple $\bar u\,{\in}\,(R_i)^\A$.
  \end{enumerate}
  The set of edges $E$ consists of
  \begin{enumerate}
    \item edges connecting the vertices \tup{u_j} and \tup{R_i,j,\bar u} if $\bar u=\tup{u_1,u_2,\ldots,u_{k_i}}$, and
    \item edges connecting the vertices \tup{R_i,j,\bar u} and \tup{R_i,j+1,\bar u} for $1\leq j<k_i$.
  \end{enumerate}
  The object graph $\OG(s)$ for a planning state $s$ is the object graph
  $\OG(\A^s)$ of its relational structure.
\end{definition}

The vertices of the form \tup{u} are called \emph{object vertices}, and vertices of the form
\tup{R,j,\bar u} are called \emph{positional-argument vertices}. The first type of
edge connects object vertices to corresponding positional-argument vertices, while
the second connects successive positional-argument vertices.

\begin{figure}[t]
  \centering
  \resizebox{.99\linewidth}{!}{
  \begin{tikzpicture}[>=latex,
                      circleNode/.style={circle, draw, minimum size=2.5em}, 
                      emptyNode/.style={},
                      edgeStyle/.style={thick},
                      rectangleNode/.style={rectangle, rounded corners=2mm, draw, minimum size=2.5em}] 
    \node[emptyNode, fill=butter1]                                    (L-gripper)          {\tup{\mathit{gripper},1,\tup{\object{L}}}};
    \node[emptyNode, fill=butter1, right=0.2cm of L-gripper]       (R-gripper)          {\tup{\mathit{gripper},1,\tup{\object{R}}}};
    \node[emptyNode, fill=aluminium2, right=0.2cm of R-gripper]            (ball)          {\tup{\mathit{ball},1,\tup{\object{b}}}};
    \node[emptyNode, below=0.3cm of L-gripper]               (L)          {\tup{\object{L}}};
    \node[emptyNode, below=0.3cm of R-gripper]               (R)          {\tup{\object{R}}};
    \node[emptyNode, below=0.3cm of ball]                    (b)          {\tup{\object{b}}};
    \draw[edgeStyle] (L-gripper) -- (L);
    \draw[edgeStyle] (R-gripper) -- (R);
    \draw[edgeStyle]      (ball) -- (b);

    \node[emptyNode, fill=chocolate1, below right=0.3cm and -0.2cm of b]   (atg1)          {\tup{\mathit{at}_g,1,\tup{\object{b},\object{B}}}};
    \node[emptyNode, fill=chameleon1, left=0.2cm of atg1]                (carry1)          {\tup{\mathit{carry},1,\tup{\object{b},\object{R}}}};
    \node[emptyNode, fill=chameleon3, left=0.3cm of carry1]              (carry2)          {\tup{\mathit{carry},2,\tup{\object{b},\object{R}}}};
    \draw[edgeStyle]   (atg1) -- (b);
    \draw[edgeStyle] (carry1) -- (b);
    \draw[edgeStyle] (carry1) -- (carry2);
    \draw[edgeStyle] (carry2) -- (R);

    \node[emptyNode, fill=chocolate2, below=0.3cm of atg1]                 (atg2)          {\tup{\mathit{at}_g,2,\tup{\object{b},\object{B}}}};
    \node[emptyNode, fill=skyblue1, left=0.2cm of atg2]                (B-room)          {\tup{\mathit{room},1,\tup{\object{B}}}};
    \node[emptyNode, fill=plum1, left=0.2cm of B-room]                  (at)          {\tup{\mathit{at}\text{-}\mathit{robot},1,\tup{\object{B}}}};
    \node[emptyNode, fill=skyblue1, left=0.2cm of at]                  (A-room)          {\tup{\mathit{room},1,\tup{\object{A}}}};
    \node[emptyNode, below=0.3cm of A-room]                  (A)          {\tup{\object{A}}};
    \node[emptyNode, below=0.3cm of B-room]                  (B)          {\tup{\object{B}}};
    \draw[edgeStyle]   (atg1) -- (atg2);
    \draw[edgeStyle]   (atg2) -- (B);
    \draw[edgeStyle] (B-room) -- (B);
    \draw[edgeStyle]     (at) -- (B);
    \draw[edgeStyle] (A-room) -- (A);
  \end{tikzpicture}
  }
  \caption{Object graph $\OG(s)$ for a state $s$ in a Gripper instance with
    grippers \object{L} and \object{R}, one ball \object{b}, and
    two rooms \object{A} and \object{B}.
    In the state $s$, the robot is at \object{B}, the ball is at gripper \object{R},
    and the goal is for the ball to be in room \object{B}.
    The state specifies the goal using the goal predicate $\mathit{at}_g$.
    This graph is isomorphic to the graph $\OG(t)$ for a state $t$ that is like $s$
    except that the ball is at gripper \object{L}.
  }
  \label{fig:object_graph}
\end{figure}
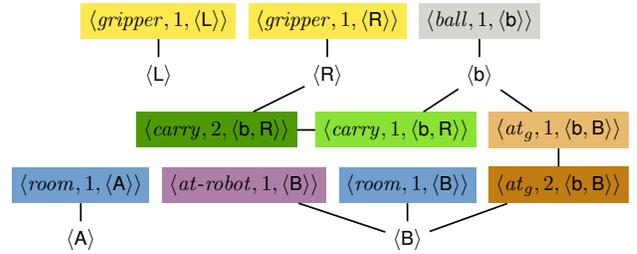

\begin{example}
  Figure~\ref{fig:object_graph} shows the object graph $\OG(s)$ for a state $s$
  of Gripper where there is a single ball, the robot is at room \object{B}, and
  the ball is being held.
  This graph is isomorphic to $\OG(t)$ where the state $t$ is like $s$, except
  that the other gripper holds the ball. \qed
\end{example}

The mapping from relation structures (states) into object graphs preserves all the
information in the structures:

\begin{theorem}[Reductions]
  \label{thm:reduction}
  Let $\A$ and $\B$ be two relational structures over a common universe $U$
  and signature (with no constant symbols).
  Then, $\A\simeq\B$ iff $\OG(\A)\simeq_g \OG(\B)$.
\end{theorem}
\Proof[Proof (sketch)]{
  First assume $\A\,{\simeq}\,\B$ with $\sigma\,{:}\,\A\,{\rightarrow}\,\B$.
  We construct a color-preserving isomorphism $f$ from $\OG(\A)$ to $\OG(\B)$:
  for object vertices, $f(\tup{u})\doteq\tup{\sigma(u)}$, while for
  positional-argument vertices, $f(\tup{R,j,\bar u})\doteq\tup{R,j,\sigma(\bar u)}$.
  It can be seen that $f$ is an edge-preserving  bijection between the vertices of
  both graphs.
  Additionally, $\lambda(\tup{u})=\bot=\lambda(\tup{\sigma(u)})$,
  and $\lambda(\tup{R,j,\bar u})=\tup{R,j}=\lambda(\tup{R,j,\sigma(\bar u)})$.
  Hence, $f$ is a color-preserving isomorphism.

  For the converse, let us assume that $f$ is a color-preserving isomorphism from $\OG(\A)$ to $\OG(\B)$.
  Consider the function $\sigma:U\,{\rightarrow}\,U$ defined by $\sigma(u)\myeq u'$ iff
  $f(\tup{u})\myeq\tup{u'}$.
  As no object vertex has the color of a positional-argument vertex,
  $\sigma$ is a $U$-permutation.
  We need to show $\sigma:\A\rightarrow\B$; i.e., for each relation $R$,
  \begin{equation}
    \label{eq}
    R^\B\ =\ \set{\sigma(\bar u)\mid \bar u\in R^\A}\,.
  \end{equation}
  The set of vertices related to the tuple $\bar u$ in $R^\A$
  is $V(\A,\bar u)\myeq\set{\tup{u_i}\mid u_i\in\bar u}\cup\set{\tup{R,j,\bar u}\mid 1\leq j\leq k}$.
  This set induces the subgraph $G(\A,\bar u)$ of $\OG(\A)$.
  It is not hard to see that \eqref{eq} holds iff the subgraphs
  $G(\A,\bar u)$ and $G(\B,\sigma(\bar u))$, for all tuples $\bar u\in R^\A$,
  are isomorphic through the (restriction of) $f$.
  As this is the case, \eqref{eq} holds, and $\A\simeq\B$.
}

By Theorem~\ref{thm:reduction}, we can use \nauty to identify equivalent states.
Other state encodings have been proposed that are not aimed at testing structural
equivalence but at using standard GNN libraries \cite{stahlberg-et-al-icaps2022,chen-et-al-nips2023wsprl-a}.
While the theoretical relationship between GNNs and first-order logics with counting
quantifiers $\C_k$ is known \cite{grohe:gnn}, the relation between logical entailment
of such logics over relational structures (i.e., states) and their different encodings
(e.g., object graphs) is not clear.
\section{Abstractions and Domain Expressivity}

Function-based policies, as defined above, such as those captured
by GNNs, do not distinguish isomorphic states.
On the other hand, such policies often need to distinguish
\emph{non-isomorphic} states as they may require different actions.

We focus on two key aspects: whether a pair of non-isomorphic states
$(s,s')$ can be distinguished with GNNs, and whether a pair of states
$(s,s')$ with different $V^*$-value can be distinguished with GNNs.
Such pairs  that cannot be distinguished by any GNN are called 
\textbf{conflict pairs.} If a training set contains conflict pairs of the first type
and $s$ is a goal state and $s'$ is not, then no  GNN will be able to
distinguish goal states from non-goal states. If the conflict is of the second type,
no GNN will learn a representation of $V^*$, even in the training set.

\Omit{
it means that there is a potential lack of expressivity, that in the extreme
case, may indicate the non-existence of general policies on a given
domain; e.g., if $(s,s')$ is a conflict pair of the first type, and
$s$ (resp.\ $s')$ is a goal (resp.\ non-goal) state, any GNN-based
general policy cannot identify when a goal state has been reached.
If the training set contains conflicts of the second type, then no
GNN-based representation of the $V^*$-function exists, which means
that approaches for generalized planning based on a GNN-based
representation of $V^*$, such as \cite{stahlberg-et-al-icaps2022}, will fail.
}

We use the known relations between the counting logics $\C_k$ and
Weisfeiler-Leman coloring algorithms \cite{cai-furer-immerman-combinatorica1992},
and the latter and GNNs \cite{morris-et-al-aaai2019,xu-et-al-iclr2019,barcelo-et-al-iclr2020,grohe:gnn},
to establish whether a domain contains conflict pairs. 
More precisely, we use the 1-WL and 2-FWL coloring algorithms
over to the object graph $G(s)$  associated with relational structures (states) $s$. 


It is known that if $s$ and $s'$ are two states whose object
graphs cannot be distinguished by 1-WL, they will not be distinguished
either by formulas in the logic $\C_2$ (first-order logic with counting quantifiers and two variables),
or by the embeddings produced by a GNN. And if the graphs
for $s$ and $s'$ cannot be distinguished by 2-FWL, they cannot be
distinguished by formulas in the logic $\C_3$ or by the
embeddings produced by 3-GNNs. 

Graphs are compared in  terms of their \emph{histograms of colors,}
denoted by $\hist^k(\cdot)$ with $k=1$ for 1-WL, and $k>1$ for $k$-FWL,
where such histogram is just the \emph{multiset of colors} for the vertices in the graph.
Namely, two states $s$ and $s'$ are distinguished if $\hist^k(G) \neq \hist^k(G')$,
where $G\myeq G(s)$ and $G'\myeq G(s')$ are the corresponding object graphs.

In the experiments, we obtain the histograms by running 1-WL and 2-FWL over
the object graphs for hundreds of training instances of different planning domains.
Let $D$ be a STRIPS planning domain, and let $\Q$ be a collection of
instances $P$ over $D$.
If $\S$ denotes the set of reachable states across the instances in $\Q$,
we want to check whether there is a pair of states $(s,s')$ in $\mathcal{S}$ that
is in \emph{conflict}  with respect to a coloring algorithm. Formally,

\begin{definition}[Conflicts]
  \label{def:conflicts}
  Let $\S$ be a set of reachable states for instances over a common domain,
  where the states are assumed to contain goal atoms.
  Further, let us consider a coloring algorithm $\times$, such as 1-WL
  (color refinement), that operates on the \textbf{object graphs} $\OG(s)$,
  and let $(s,s')$ be a pair of states in $\S$ that have the \textbf{same color histogram;}
  i.e., $\hist^\times(s)\,{=}\,\hist^\times(s')$. Then,
  \begin{enumerate}[1.]
    \item $(s,s')$ is an \textbf{E-conflict} if $s \not\iso s'$, and
    \item $(s,s')$ is a \textbf{V-conflict} if $V^*(s)\neq V^*(s')$.
  \end{enumerate}
  We say that $\S$ has no conflicts of some type iff there is no pair
  $(s,s')$ in $\S$ that is a conflict of such type.
\end{definition}

Conflicts of the first type imply that GNNs cannot distinguish some pairs of
non-isomorphic states, while conflicts of the second type imply that GNNs cannot
distinguish some pairs of states that have different costs.
The proof of the following theorem follows directly from the known
correspondences between 1-WL and GNNs:

\begin{theorem}[GNN-based Representation of $V^*$]
  \label{thm:conflicts}
  Let $\Q$ be a \textbf{finite} class of problems over a common domain $D$
  (where states encode goals with goal atoms), and let $\S$ be the set of
  reachable states in $\Q$. Then,
  \begin{enumerate}[1.]
    \item $\mathcal{S}$ has no E-conflicts of type 1-WL iff there is a GNN
      that identifies the states $[s]$ in the abstraction $\isoreduct{\Q}$, and
    \item $\mathcal{S}$ has no V-conflicts of type 1-WL iff there is a GNN
      that represents the value function $V^*(s)$ over $\S$.
  \end{enumerate}
\end{theorem}

\Omit{
\alert{%
  \begin{enumerate}[1.]
    \item Not clear what to do about Theorem about objects as it seems difficult
      to include it here.
    \item One possibility: Let $s$ be a state, and let $C_1$ and $C_2$ be two sets
      of objects of the equal cardinality, such that all objects in $C_1$ (resp.\ $C_2$)
      have the same 1-WL color in $s$.
      Let $s_1$ and $s_2$ be two states that extend $s$ with $p$-atoms that establish
      two different 1-to-1 relation between $C_1$ and $C_2$.
      Then, $(s_1,s_2)$ is an E-conflict, and moreover, if they have different $V^*$-value,
      then $(s_1,s_2)$ is a V-conflict.
    \item The problem with this is that I'm not sure it is true. We know that the equitable
      partition for $s$ is equal to the equitable partitions for $s_1$ and $s_2$,
      but not sure that this implies that the 1-WL coloring histogram of $s_1$ is equal
      to that of $s_2$. Indeed, I think they may be different.
  \end{enumerate}
}
}


\section{Experiments: Domain Expressivity}

Experiments are carried out to evaluate the expressivity requirement
of various planning domains by looking for E- and V-conflicts.
Testing for the equivalence relation $\iso$ is implemented in Python
using the planning library Mimir ~\cite{stahlberg-ecai2023} and \nauty,
while for computing color histograms we implemented 1-WL and 2-FWL \cite{drexler-et-al-zenodo2024b}.
The benchmark set consists of domain and instances from the
International Planning Competition (IPC). Code and data  are available online \cite{drexler-et-al-zenodo2024b}.

\newcommand{\X}{\cellcolor{orange1}}
\newcommand{\Y}{\cellcolor{chameleon1}}

\setlength{\tabcolsep}{3.4pt}
\begin{table*}[ht]
  \centering
  \resizebox{\linewidth}{!}{
  \begin{tabular}{@{}l rrr rrrrrrrr rrrrrrrr@{}}
    \toprule
                     & & & & \multicolumn{8}{c}{Multisets} & \multicolumn{8}{c}{Standard sets} \\ \cmidrule(lr){5-12} \cmidrule(l){13-20}
                     & & & & \multicolumn{2}{c}{1-WL} & \multicolumn{2}{c}{2-FWL} & \multicolumn{2}{c}{1-WL + G} & \multicolumn{2}{c}{2-FWL + G} & \multicolumn{2}{c}{1-WL} & \multicolumn{2}{c}{2-FWL} & \multicolumn{2}{c}{1-WL + G} & \multicolumn{2}{c}{2-FWL + G} \\ \cmidrule(lr){5-6} \cmidrule(lr){7-8} \cmidrule(lr){9-10} \cmidrule(lr){11-12} \cmidrule(lr){13-14} \cmidrule(lr){15-16} \cmidrule(lr){17-18} \cmidrule(l){19-20}
    Domain           & $\#\Q$  & $\#\S$ & $\#\isoreduct{\S}$ & $\#E$ & $\#V$ & $\#E$ & $\#V$ & $\#E$ & $\#V$ & $\#E$ & $\#V$ &  $\#E$ & $\#V$ & $\#E$ & $\#V$ & $\#E$ & $\#V$ & $\#E$ & $\#V$ \\ \midrule
    Barman           &    510  &  115~M &               38~M & 1,326 &   537 &     0 &     0 &\Y1,062 &\Y273 &     0 &     0 &  1,326 &   537 &     0 &     0 &\Y1,062 &\Y273 &     0 &     0 \\
    Blocks3ops       &    600  &  146~K &              133~K &    50 &    20 &     0 &     0 &\Y  25 &\Y   0 &     0 &     0 &     50 &    20 &     0 &     0 &\Y  25 &\Y   0 &     0 &     0 \\
    Blocks4ops       &    600  &  122~K &              110~K &    54 &    27 &     0 &     0 &\Y  27 &\Y   0 &     0 &     0 &     54 &    27 &     0 &     0 &\Y  27 &\Y   0 &     0 &     0 \\
    Blocks4ops-clear &    120  &   31~K &                3~K &     0 &     0 &     0 &     0 &     0 &     0 &     0 &     0 &      0 &     0 &     0 &     0 &     0 &     0 &     0 &     0 \\
    Blocks4ops-on    &    150  &   31~K &                8~K &     0 &     0 &     0 &     0 &     0 &     0 &     0 &     0 &      0 &     0 &     0 &     0 &     0 &     0 &     0 &     0 \\
    Childsnack       &     30  &   58~K &                5~K &     0 &     0 &     0 &     0 &     0 &     0 &     0 &     0 &      0 &     0 &     0 &     0 &     0 &     0 &     0 &     0 \\
    Delivery         &    540  &  412~K &               62~K &     0 &     0 &     0 &     0 &     0 &     0 &     0 &     0 &\X  152 &     0 &     0 &     0 &\X 152 &     0 &     0 &     0 \\
    Ferry            &    180  &    8~K &                4~K &    36 &    36 &     0 &     0 &\Y   0 &\Y   0 &     0 &     0 &\X   84 &\X  84 &     0 &     0 &\Y   0 &\Y   0 &     0 &     0 \\
    Grid             &  1,799  &  438~K &              370~K &    42 &    38 &     0 &     0 &\Y  24 &\Y  20 &     0 &     0 &\X   84 &\X  80 &     0 &     0 &\Y  44 &\Y  40 &     0 &     0 \\
    Gripper          &      5  &    1~K &                 90 &     0 &     0 &     0 &     0 &     0 &     0 &     0 &     0 &      0 &     0 &     0 &     0 &     0 &     0 &     0 &     0 \\
    Hiking           &    720  &   44~M &                5~M &     0 &     0 &     0 &     0 &     0 &     0 &     0 &     0 &      0 &     0 &     0 &     0 &     0 &     0 &     0 &     0 \\
    Logistics        &    720  &   69~K &               38~K &   131 &   131 &     0 &     0 &\Y  94 &\Y  94 &     0 &     0 &    131 &   131 &     0 &     0 &\Y  94 &\Y  94 &     0 &     0 \\
    Miconic          &    360  &   32~K &               22~K &     0 &     0 &     0 &     0 &     0 &     0 &     0 &     0 &      0 &     0 &     0 &     0 &     0 &     0 &     0 &     0 \\
    Reward           &    240  &   14~K &               11~K &     0 &     0 &     0 &     0 &     0 &     0 &     0 &     0 &      0 &     0 &     0 &     0 &     0 &     0 &     0 &     0 \\
    Rovers           &    514  &   39~M &               34~M &     0 &     0 &     0 &     0 &     0 &     0 &     0 &     0 &      0 &     0 &     0 &     0 &     0 &     0 &     0 &     0 \\
    Satellite        &    960  &   14~M &                8~M & 5,304 & 4,226 &     0 &     0 &\Y1,708 &\Y762 &     0 &     0 &\X12,908 &\X9,906 &   0 &     0 &\Y4,372 &\Y982 &     0 &     0 \\
    Spanner          &    270  &    9~K &                4~K &     0 &     0 &     0 &     0 &     0 &     0 &     0 &     0 &      0 &     0 &     0 &     0 &     0 &     0 &     0 &     0 \\
    Visitall         &    660  &    3~M &                2~M &     0 &     0 &     0 &     0 &     0 &     0 &     0 &     0 &\X   27 &     0 &     0 &     0 &\X  27 &     0 &     0 &     0 \\
    \bottomrule
  \end{tabular}}
  \caption{%
    The column $\#\Q$ is the number of instances used in our experiments.
    The columns denoted $\#\S$ and $\#\isoreduct{\S}$ refer to the the total number of states and the total number of partitions in the expanded reachable state spaces.
    The left part uses a multiset, while the right part uses  sets.
    The suffix "G" indicates that goal atoms are marked as true if they hold true in the state.
    The number of conflicts that are caused by $1$-WL and $2$-FWL where
    the $\#E$ column refers to the total number of conflicts, while the $\#V$ column refers to the number of conflicts in which the two classes differ in V$^*$.
  }
  \label{table:conflicts}
\end{table*}

Conflicts are calculated with respect to 1-WL and 2-FWL, and also versions of
these algorithms in which multisets are replaced by standard sets.\footnote{Coloring
  algorithms work with multisets, rather than sets, as multisets provide a  means
  to do restricted forms of counting.
}
This modification is important because the description logic grammar that
is used to generate state features from the planning domain does not use
counting quantifiers, and also because some GNN-based approaches use
max-aggregation rather than sum-aggregation
\cite{stahlberg-et-al-icaps2022,stahlberg-et-al-kr2022,stahlberg-et-al-kr2023}.

Table~\ref{table:conflicts} shows the number of E- and V-conflicts
among the reachable states in the benchmark.
The table shows, for each domain, the number of instances and their reachable
states ($\#\Q$ and $\#\S$), the number of equivalence classes
($\#\isoreduct{\S}$), and the number of E- and V-conflicts ($\#E$ and $\#V$,
respectively) for 1-WL and 2-FWL, and for the two versions of the
algorithms (multisets and standard sets).

We also tried a slightly different graph encoding to overcome some of the limitations
of object graphs in relation to the coloring algorithms.
In this encoding, goals are represented using two predicates, $p_{g,T}$ and $p_{g,F}$,
rather than a single predicate $p_g$, that tell whether the goal atom is true or false in the state.
This encoding, called \emph{goal marking,} is beneficial in all domains that have conflicts,
highlighted in green in the columns ``1-WL + G'' and ``2-FWL + G'' in Table~\ref{table:conflicts}.
In Blocks, for example, $\#V$ drops to $0$, while in Ferry, it resolves all conflicts.

The existence of V-conflicts are important when learning a representation of $V^*$,
but E-conflicts give a more general
view on the expressivity requirements since V-conflicts are E-conflicts and
E-conflicts imply the existence of qualitatively different states that cannot
be differentiated.
As can be observed on Table~\ref{table:conflicts}, and by Theorem~\ref{thm:conflicts}:
\begin{enumerate}[$\bullet$]
  \item 1-WL (and hence GNNs) has sufficient expressive power in 11 domains (61\%), where there are no conflicts at all.
  \item In 12 (resp.\ 14) domains, 1-WL has sufficient expressive power to separate non-isomorphic states (resp.\ represent $V^*$)
    when using goal marking.
  \item In some domains, 1-WL is not expressive enough even with goal marking; this includes the domains Barman, Grid, Logistics, and Satellite.
  \item Most important, \emph{2-FWL, that has the expressive power of $\C_3$, appears to be sufficiently expressive in all domains.}
\end{enumerate}

The table also shows that reducing expressiveness by using sets instead of multisets
does not reduce the expressive power needed in most domains.
Indeed, the modified 1-WL algorithm with sets  creates E-conflicts in Delivery and Visitall
(highlighted in orange), but no  V-conflicts where  they were none. Indeed, it  just  increases the number of  conflicts
in Ferry, Grid, and Satellite which  was not zero with multisets. 

\citeay{stahlberg-et-al-kr2023} noted that Logistics requires $\C_3$ features to learn a value function,
and similarly for Grid \cite{stahlberg-et-al-arxiv2024}.
\emph{The experiments corroborate these claims, as 1-WL found conflicts in these domains.}
However, \citeay{stahlberg-et-al-icaps2022} claim that Rovers requires $\C_3$ features, but no conflicts are identified.
This finding does not disprove the claim because Rovers contains an important ternary predicate, \textsc{Can-Traverse};
rather, it likely  suggests that our training set is not sufficiently rich.

Barman, Ferry, and Satellite, as far as we know, have not been previously analyzed in this context.
The conflicts in Blocks have been studied by~\citeay{horcik-sir-icaps2024}, where they show that if the goal has a specific structure,
then $\C_2$ cannot determine if it is true in a state.
Logistics has been investigated by~\citeay{stahlberg-et-al-kr2023}, where they used \emph{derived predicates} to ensure $\C_2$ is sufficient to express a policy.
The results suggest that the expressiveness of $1$-WL is insufficient for learning a value function.
We now study these domains and the conflicts we have identified.

\medskip\noindent\textbf{Barman.}
The objective is to mix cocktails that require exactly $2$ ingredients.
To create the cocktails, the bartender can fill shot glasses with specific ingredients, pour the shot glasses into a shaker, mix the ingredients with the shaker, and clean the shot glasses and the shaker.
A typical plan for creating a cocktail involves pouring the first ingredient into a shot glass, transferring it to the shaker, cleaning the shot glass, pouring the second ingredient into it, then into the shaker, cleaning the shot glass again, shaking the shaker, and finally pouring the cocktail into a shot glass.
Figure~\ref{figure:barman:conflict} illustrates two states with different $V^*$ values that cannot be distinguished by $1$-WL.
There are two different cocktail recipes, $c_1$ requiring ingredients $i_1$ and $i_3$, and $c_2$ requiring ingredients $i_1$ and $i_2$.
The goal is to fill shot glass $s_1$ with $c_1$ and shot glass $s_2$ with $c_2$.
In both states, the shaker is on the table, and both shots are being held.
The distinction lies in the contents of the shot glasses.
In the first state, shot glass $s_1$ contains $i_3$ and shot glass $s_2$ contains $i_2$, while in the second state, shot glass $s_1$ contains $i_2$ and shot glass $s_2$ contains $i_3$.
In other words, the contents of the shot glasses have been swapped.
However, the goal specifies that shot glass $s_1$ must precisely contain cocktail $c_1$, so the optimal plan for the second state is first to pour out the contents of $s_1$ and then clean it,
as it contains the wrong ingredient, steps that are unnecessary for the first state.

\usetikzlibrary{patterns}

\begin{figure}
    \centering
    \resizebox{0.40\textwidth}{!}{%
        \begin{circuitikz}
            \node [] at (10,14.0) {Shot $s_1$};
            \node [] at (11.5,14) {Shot $s_2$};
            \draw[black, thick] (9.5,13.75) -- (9.5,12.25) -- (10.5,12.25) -- (10.5,13.75);
            \node [] at (10,13.4) {$i_1$};
            \node [] at (10,13.0) {$+$};
            \node [] at (10,12.6) {$i_3$};

            \draw[black, thick] (11,13.75) -- (11,12.25) -- (12,12.25) -- (12,13.75);
            \node [] at (11.5,13.4) {$i_2$};
            \node [] at (11.5,13.0) {$+$};
            \node [] at (11.5,12.6) {$i_3$};

            \node [] at (14,14.0) {Shot $s_1$};
            \node [] at (15.5,14) {Shot $s_2$};
            \draw[black, thick] (13.5,13.75) -- (13.5,12.25) -- (14.5,12.25) -- (14.5,13.75);
            \node [] at (14,13) {$i_3$};
            \draw[black, thick] (15,13.75) -- (15,12.25) -- (16,12.25) -- (16,13.75);
            \node [] at (15.5,13) {$i_2$};

            \node [] at (18,14.0) {Shot $s_1$};
            \node [] at (19.5,14) {Shot $s_2$};
            \draw[black, thick] (17.5,13.75) -- (17.5,12.25) -- (18.5,12.25) -- (18.5,13.75);
            \node [] at (18,13) {$i_2$};
            \draw[black, thick] (19,13.75) -- (19,12.25) -- (20,12.25) -- (20,13.75);
            \node [] at (19.5,13) {$i_3$};

            \node [font=\LARGE] at (16.75,13) {$\equiv$};
            \node [] at (10.75,11.9) {Goal};
            \node [] at (14.75,11.9) {$V^* = 23$ };
            \node [] at (18.75,11.9) {$V^* = 26$};
        \end{circuitikz}
    }%
    \caption{%
    Example of two Barman states with different $V^*$ value from the same instance that are considered isomorphic by $1$-WL with respect to the goal.
    The left (resp.~right) one in being held in the left (resp.~right) hand, and the shaker (omitted) is on the table.
    The goal is to have cocktail $c_1$ in shot glass $s_1$ and $c_2$ in $s_2$.
    The only difference in both states is that the ingredients in both shots are swapped.
    However, in the state on the right, the ingredient $i_2$ in $s_1$ is wrong and must be removed, resulting in different $V^*$ values.}
    \label{figure:barman:conflict}
\end{figure}
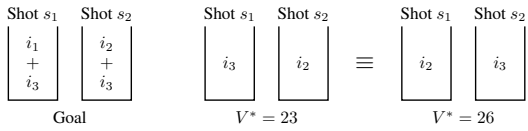

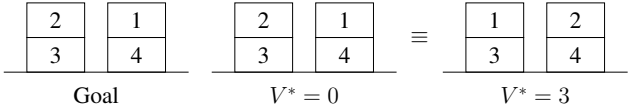
\begin{figure}[t]
    \centering
    \resizebox{0.47\textwidth}{!}{%
        \begin{circuitikz}
            \tikzstyle{every node}=[font=\LARGE]
            \draw [short] (8.25,12.25) -- (12.25,12.25);
            \draw  (8.75,13) rectangle  node {\LARGE 3} (10,12.25);
            \draw  (10.5,13) rectangle  node {\LARGE 4} (11.75,12.25);
            \draw  (8.75,13.75) rectangle  node {\LARGE 2} (10,13);
            \draw  (10.5,13.75) rectangle  node {\LARGE 1} (11.75,13);

            \draw [short] (12.75,12.25) -- (16.75,12.25);
            \draw  (13.25,13) rectangle  node {\LARGE 3} (14.5,12.25);
            \draw  (15,13) rectangle  node {\LARGE 4} (16.25,12.25);
            \draw  (13.25,13.75) rectangle  node {\LARGE 2} (14.5,13);
            \draw  (15,13.75) rectangle  node {\LARGE 1} (16.25,13);

            \draw [short] (17.75,12.25) -- (21.75,12.25);
            \draw  (18.25,13) rectangle  node {\LARGE 3} (19.5,12.25);
            \draw  (20,13) rectangle  node {\LARGE 4} (21.25,12.25);
            \draw  (18.25,13.75) rectangle  node {\LARGE 1} (19.5,13);
            \draw  (20,13.75) rectangle  node {\LARGE 2} (21.25,13);

            \node [font=\LARGE] at (17.25,13) {$\equiv$};
            \node [] at (10.25,11.75) {Goal};
            \node [] at (14.75,11.75) {$V^* = 0$};
            \node [] at (19.75,11.75) {$V^* = 3$};
        \end{circuitikz}
    }%
    \caption{%
    Example of two Blocks states that are considered isomorphic by $1$-WL with respect to the goal.
    In the object graphs, $1$-WL cannot determine whether the goal holds.}
    \label{figure:blocks:conflict}
\end{figure}

\medskip\noindent\textbf{Blocks.}
The goal is to arrange all the blocks into a specific configuration by stacking and unstacking them.
There are two versions of this domain, one with three action schemas and the other with four action schemas.
Remarkably, GNNs have been successfully trained for this domain and exhibit good generalization~\cite{stahlberg-et-al-kr2022,stahlberg-et-al-kr2023}.
However, our results, along with those of others~\cite{horcik-sir-icaps2024}, suggest that GNNs might lack the necessary expressiveness for this domain.
Figure~\ref{figure:blocks:conflict} illustrates two states and a goal description that cannot be distinguished by $1$-WL.
In this figure, the two states have distinct values: one is a goal state, and the other is not.
The object graph for the state on the left contains two connected components, each forming a $6$-gon,
and the object graph for the state on the right contains one connected component, forming a $12$-gon.
These two structures cannot be distinguished by $1$-WL.

\medskip\noindent\textbf{Ferry.}
There is only one ferry, capable of carrying a single car.
The cars can both board and disembark from the ferry, and the ferry can sail between locations.
The goal is to transport cars to their respective destinations, as denoted by a binary predicate.
The simplest states where $1$-WL fails to differentiate are those where the two cars must be in different locations.
One state has both cars at their destinations, while the other has their locations swapped.
Consequently, their values differ, with one being a goal state and the other not.
By marking goal atoms as true or false, these two states can be distinguished.

\begin{figure}
    \centering
    \scalebox{0.7}{
    \begin{tikzpicture}
        \draw (0,1) -- (0,3);
        \draw (1,0) -- (1,3);
        \draw (2,0) -- (2,3);
        \draw (3,0) -- (3,2);

        \draw (1,0) -- (3,0);
        \draw (0,1) -- (3,1);
        \draw (0,2) -- (3,2);
        \draw (0,3) -- (2,3);



        \node at (2.5,1.5) {$k_1$};
        \node at (0.5,1.5) {$k_2$};

        \node at (1.5,1.5) {$a$};

        \draw[->] (2.5,1.25) -- (2.5,0.5);
        \draw[->] (0.5,1.75) -- (0.5,2.5);

        \node [font=\Large] at (3.5,1.5) {$\equiv$};
        \node [] at (1.5,-0.4) {$V^* = 10$};
    \end{tikzpicture}
    \begin{tikzpicture}
        \draw (0,1) -- (0,3);
        \draw (1,0) -- (1,3);
        \draw (2,0) -- (2,3);
        \draw (3,0) -- (3,2);

        \draw (1,0) -- (3,0);
        \draw (0,1) -- (3,1);
        \draw (0,2) -- (3,2);
        \draw (0,3) -- (2,3);



        \node at (0.5,1.5) {$k_1$};
        \node at (2.5,1.5) {$k_2$};

        \node at (1.5,1.5) {$a$};

        \draw[->] (0.75,1.5) -- (2.5,0.5);
        \draw[->] (2.25,1.5) -- (0.5,2.5);

        \node [] at (1.5,-0.4) {$V^* = 12$};
    \end{tikzpicture}
    }
    \caption{%
    An example of two Grid states that are considered isomorphic by the $1$-WL algorithm with respect to the goal.
    The goal is to move the keys $k_1$ and $k_2$ to specific cells, as the arrows indicate.
    All keys and locks have the same shape. The agent $a$ is in the center of the grid.
    In the left state, $10$ actions are needed to solve the instance, while $12$ actions are needed in the right state.
    }
    \label{figure:grid:conflict}
\end{figure}
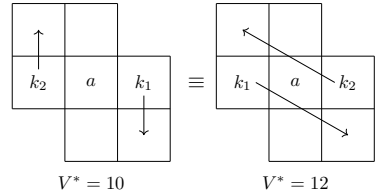

\medskip\noindent\textbf{Grid.}
In this domain, an agent needs to move keys to specific cells by picking them up and placing them down.
However, there are locked doors, and the cells might be positioned behind one.
Each locked door can only be opened by keys with the corresponding shape, i.e., both locks and keys have shapes associated with them.
An example illustrating when $1$-WL is insufficient for distinguishing non-isomorphic states is shown in Figure~\ref{figure:grid:conflict}.
In these states, the positions of two keys have been swapped, resulting in different $V^*$ values.
However, $1$-WL cannot determine which key should be placed in which location.

\newcommand{\Tpre}{\ensuremath{\text{T}_\text{pre}}}
\newcommand{\Tlearn}{\ensuremath{\text{T}_\text{learn}}}

\FPeval{\suBlocksA}{round((992+28781)/(213+11020),2)}
\FPeval{\suBlocksB}{round((3+5)/(3+3),2)}
\FPeval{\suBlocksC}{round((30+77)/(33+195),2)}
\FPeval{\suDelivery}{round((107+427)/(65+260),2)}
\FPeval{\suFerry}{round((13+56)/(19+72),2)}
\FPeval{\suGripper}{round((2+3)/(2+4),2)}
\FPeval{\suMiconic}{round((8+30)/(14+44),2)}
\FPeval{\suReward}{round((5+15)/(6+8),2)}
\FPeval{\suSpanner}{round((3+4)/(4+4),2)}
\FPeval{\suVisitall}{round((46+15487)/(1761+14163),2)}
\FPeval{\faBlocksA}{round(145680/4901,2)}
\FPeval{\faBlocksB}{round(30540/86,2)}
\FPeval{\faBlocksC}{round(30540/249,2)}
\FPeval{\faDelivery}{round(411720/3346,2)}
\FPeval{\faFerry}{round(8430/265,2)}
\FPeval{\faGripper}{round(1084/90,2)}
\FPeval{\faMiconic}{round(32400/12339,2)}
\FPeval{\faReward}{round(13394/7026,2)}
\FPeval{\faSpanner}{round(9291/283,2)}
\FPeval{\faVisitall}{round(476766/402880,2)}

\begin{table*}
  \centering
  \footnotesize
  \begin{tabular}{@{} l c rrrr c rrrrrr @{}}
    \toprule
    &\textcolor{white}{x}&\multicolumn{4}{c@{}}{without equivalence-based reduction} &\textcolor{white}{xx}& \multicolumn{6}{c@{}}{with equivalence-based reduction} \\
    \cmidrule{3-6}
    \cmidrule{8-13}
    Domain              && M & \Tpre & \Tlearn & $\#\QT$  &&    M &  \Tpre &   \Tlearn & Speedup       & $\#\isoreduct{\QT}$ & Factor      \\
    \cmidrule{1-1}
    \cmidrule{3-6}
    \cmidrule{8-13}
    Blocks3ops          &&\bf 9 &\bf 103 &  28,781 & 145,680  &&   11 &    213 &\bf 11,020 &\bf\suBlocksA  &\bf            4,901 & \faBlocksA  \\
    Blocks4ops-clear    &&    1 &      3 &       5 &  30,540  &&    1 &      3 &\bf      3 &\bf\suBlocksB  &\bf               86 & \faBlocksB  \\
    Blocks4ops-on       &&    3 &\bf  30 &\bf  177 &  30,540  &&\bf 2 &     33 &       195 &   \suBlocksC  &\bf              249 & \faBlocksC  \\
    Delivery            &&    3 &    107 &     427 & 411,720  &&\bf 2 &\bf  65 &\bf    260 &\bf\suDelivery &\bf            3,346 & \faDelivery \\
    Ferry               &&    1 &\bf  13 &\bf   56 &   8,430  &&    1 &     19 &        72 &   \suFerry    &\bf              265 & \faFerry    \\
    Gripper             &&    1 &      2 &\bf    3 &   1,084  &&    1 &      2 &         4 &   \suGripper  &\bf               90 & \faGripper  \\
    Miconic             &&    1 &\bf   8 &\bf   30 &  32,400  &&    1 &     14 &        44 &   \suMiconic  &\bf           12,339 & \faMiconic  \\
    Reward              &&    1 &\bf   5 &      15 &  13,394  &&    1 &      6 &\bf      8 &\bf\suReward   &\bf            7,026 & \faReward   \\
    Spanner             &&    1 &\bf   3 &       4 &   9,291  &&    1 &      4 &         4 &   \suSpanner  &\bf              283 & \faSpanner  \\
    Visitall            &&\bf 2 &\bf  22 &\bf   55 & 476,766  &&    3 &     36 &        59 &   \suVisitall &\bf          402,880 & \faVisitall \\
    \bottomrule
  \end{tabular}
  \caption{%
    Learning general policies with and without equivalence-based reductions.
    The table shows the memory in GiB (M), the wall-clock times in seconds for preprocessing ($\text{T}_\text{pre}$),
    the time in seconds for grounding, solving the ASPs, and validation ($\text{T}_\text{learn}$), the total number
    of states in the training set ($\#\QT$), and the reduced training set ($\#\isoreduct{\QT}$), and ratios for the
    speedup in time and number of states for the reduced training set.
    Boldface figures denote the winner in the pairwise comparison, i.e., the one with strictly fewer resources needed.
  }
  \label{table:learning}
\end{table*}

\medskip\noindent\textbf{Logistics.}
This domain involves cities, trucks, airplanes, and packages.
In each city, there are several locations where trucks can move between, as well as pick up and deliver packages.
There is also an airport in each city, from which airplanes can load and unload packages.
The goal is to deliver each package to a specific location within some city.
A plan for a single package typically involves a truck that picks it up and unloads it at the airport, then an airplane is used to move it to the correct city, after which a truck is used to deliver it to the destination.
Two states with different $V^*$ values that $1$-WL cannot discriminate are as follows:
There are two cities, $c_1$ and $c_2$, each consisting of a single location, which we refer to using the city name.
There is a single truck in each location, $t_1$ at $c_1$ and $t_2$ at $c_2$.
There are also two airplanes, $a_1$ at $c_1$ and $a_2$ at $c_2$.
The goal is to deliver two packages, $p_1$ to $c_1$ and $p_2$ to $c_2$.
In one state, $p_1$ is inside $t_1$ and $p_2$ is inside $t_2$, while in the other state, $p_1$ is inside $t_2$ and $p_2$ is inside $t_1$.
The $V^*$ value of the first state is $2$ as the trucks have to unload the packages, while the value is $8$ in the second state as they need to be transported to the other city.
Here, $1$-WL is unable to determine whether the correct packages are inside the trucks.

\medskip\noindent\textbf{Satellite.}
In this domain, there are satellites equipped with instruments to capture specific images.
Each satellite can calibrate the equipment to various targets, but not necessarily to all possible targets.
The typical goal is to capture images of various phenomena using specific instruments.
We found states with different values that are identified as isomorphic by $1$-WL.
One example is an instance where the goal is to capture a spectrograph image of a phenomenon, and there are two satellites capable of capturing such an image.
However, only one satellite can calibrate the instrument to the phenomenon; thus, said satellite has to capture the image.
The only difference between the two states is that, in the first state, one satellite is pointing to the ground station and the other is pointing to a star related to the phenomenon, whereas in the second state, their orientations have been swapped.
This means that in one state, one satellite must first turn to the star to calibrate the instrument.
However, $1$-WL is unable to determine whether the correct satellite points to the star -- only that one satellite does.


\section{Experiments: Learning on Abstractions}

The next set of experiments evaluates the impact of replacing the states in the training
set when learning general policies with symbolic methods \cite{drexler-et-al-icaps2022}
with their abstractions.
For both training sets, the learned policies are aimed to generalize to a much larger
(infinite) class of instances.
The impact on performance for symbolic learning mainly results from reducing in
the number of states, although some extra preprocessing is needed to implement the
reduction, which, for the easiest cases, increases overall times.
If $\Q_T$  denotes the set of states used for training, then \isoreduct{\Q_T} denotes
the reduced set of states obtained in the equivalence-based abstraction where every
pair of isomorphic states in $\Q_T$ are mapped to the same abstract state.

Learning is done on two Intel Xeon Gold 6130 CPUs with 32 cores,
96 GiB of memory, and a time budget of 24 hours.
Since the reductions are significant, we use training instances with up to 10,000
states instead of the 2,000 used by~\citeay{drexler-et-al-icaps2022}, and we tested
generalization of the learned policies on significantly larger instances.

Table~\ref{table:learning} shows a summary of the times required for preprocessing
(that includes the tests for $\iso$) and the learning of the general policies.
The sizes of the plain and reduced training sets, $\#\QT$ and $\#\isoreduct{\QT}$
respectively, are shown, as well as the reduction factors with respect to
time (Speedup) and the number of states (Factor).
Notice that there is only a single state in \isoreduct{\QT} for every equivalence class across all instances.
As it can be seen, the total overhead incurred by testing $\iso$ (i.e., the difference
between the two figures for \Tpre) is small.

Policy learning is done iteratively by solving a Clingo program (ASP) over a
\emph{subset} of the training set that is grown at each iteration until the
resulting policy correctly solves (i.e., verifies) \emph{all} the instances in
the training set.
Table~\ref{table:learning} shows that the learning time increases
for the easiest cases due to the overhead but reduces for the most difficult domains, Blocks3ops and Delivery.
Our policy learning code is not optimized as it is implemented on top of the code
for learning sketches \cite{drexler-et-al-icaps2022}, a task that requires further bookkeeping.
We expect better speedups by using specific code only for policy learning
because they do not require computing the complete abstraction mapping and,
therefore, can better exploit the reduction in abstract states.

\section{Discussion}

In recent work, developed independently,  \citeay{horcik-sir-icaps2024}
analyze the expressive power of a number of GNN architectures over a number of
planning domains. For this, they map state pairs $s$ and $s'$ from a domain
instance into  graphs, and run  GNNs with random weights to compute scalars
$g(s)$ and $g(s')$.\footnote{Other mappings from states into graphs are
  considered by \citeay{chen-et-al-nips2023wsprl-a} and \citeay{chen-et-al-nips2023wsprl-b}.
}
The equality $g(s) = g(s')$ is a strong indication that the GNNs
cannot distinguish $s$ from $s$', and if the
actual costs $V^*(s)$ and  $V^*(s')$ are different,
the pair  $(s,s')$ is marked as a conflict;
an  indication that  GNNs lack expressive
power to capture $V^*$ in  the domain.
In our case, rather than using GNNs with random weights,
we run 1-WL, and rather than using different types of
graphs, we use a map from states (relational structures)
to graphs that is invariant under state isomorphism.
In addition, we see if 1-WL distinguishes
non-isomorphic pairs of states and not
just states with different $V^*$ values.
This is important because E-conflicts
$(s,s')$, as we call them, may
become V-conflicts when the goals
encoded in $s$ and $s'$ change.
Yet, while  results over the various domains
are quite different, the reasons for these
differences may be elsewhere.
\citeay{horcik-sir-icaps2024} consider large training instances
but sample the state pairs that are considered;
we consider small training instances
and consider all possible state pairs.
The result is that we observe conflicts
in domains such as Barman, Blocks, Logistics,
and Satellite, but not in Rovers,
while they observe conflicts in Rovers
but not in the first four domains.

While the presence of V-conflicts in a domain is a strong indication that
GNNs will not be able to represent the optimal value function, even over the training
instances, the lack of V-conflicts does not ensure that the GNNs will represent the
optimal value function or suitable approximation of it over the test set (as in Rovers).
Also, GNNs may fail to represent $V^*$ over the training set and yet accommodate
non-optimal policies.  Likewise, in certain cases, this limitation can be addressed
by using slightly different  state encodings, as shown in the case of  Blocks and Ferry
where  goal and state predicates $p_g$ and $p$ are composed. Other ways
for extending the state representations are addressed by
\citeay{stahlberg-et-al-arxiv2024}.
\section{Conclusions}

State symmetries play two key roles in generalized planning. On the one hand, symmetric  states
can be pruned, speeding up the learning process with no information loss. On the other hand, non-symmetric states
need to be distinguished by the languages and neural architectures used to represent and learn value functions and policies.
Indeed, languages and architectures that lack the  expressive power to make these distinctions may fail to accommodate general
policies for certain planning domains at all. These two  roles of symmetries and non-symmetries have 
been studied through a number of experiments that illustrate the expressive power required by some common planning domains
and the performance gains  obtained in  the symbolic setting for learning general policies. 
In the future, we want to explore how these results can be sharpened and made more broadly useful by learning
general policies for domains that remain out of reach for current techniques.

\section*{Acknowledgments}

This work has been supported by the Alexander von Humboldt Foundation with funds from the Federal Ministry for
Education and Research. It has also received funding from the European Research Council (ERC), Grant agreement No
885107, the Excellence Strategy of the Federal Government and the NRW Lander, Germany, and the Knut and Alice
Wallenberg (KAW) Foundation under the WASP program. The computations were enabled in part  by the supercomputing
resource Berzelius provided by National Supercomputer Centre at Link\"{o}ping University and the KAW Foundation.

\bibliographystyle{kr}
\bibliography{../../bib/abbrv-short,../../bib/abbrv,../../bib/literatur,extra,../../bib/crossref-short}

\end{document}